\documentclass[runningheads]{llncs}
\usepackage{graphicx}
\usepackage[]{algorithm2e}
\begin{document}
\title{Data-driven PDE discovery with evolutionary approach}
%
%
\author{Michail Maslyaev \and
Alexander Hvatov \and Anna Kalyuzhnaya }
\authorrunning{M. Maslyaev et al.}
%
\institute{ITMO University, Kronsersky pr. 49, 197101, St. Petersburg, Russia
\email{alex\_hvatov@corp.ifmo.ru}}

\maketitle              
\begin{abstract}
The data-driven models allow one to deﬁne the model structure in cases when a priori information is not suﬃcient to build other types of models. The possible way to obtain physical interpretation is the data-driven differential equation discovery techniques. The existing methods of PDE (partial derivative equations) discovery are bound with the sparse regression. However, sparse regression is restricting the resulting model form, since the terms for PDE are defined before regression. The evolutionary approach described in the article has a symbolic regression as the background instead and thus has fewer restrictions on the PDE form. The evolutionary method of PDE discovery (EPDE) is described and tested on several canonical PDEs. The question of robustness is examined on a noised data example.

\end{abstract}
\section{Introduction}

Data-driven algorithms are usually considered as the source of models when the connection between the data samples is not known a priori. There are various data-driven models built on neural networks, regression, statistic, etc. As an example, deep neural networks models \cite{DLmod}, recurrent-convolutional neural network models \cite{RNmod}, statistical models \cite{nikitin2016statistics}, regression \cite{KIM20181736}, combined evolutionary-based models \cite{kovalchuk2018conceptual} and other models and their combinations \cite{gusarov2017spatially}. However, most of the existing models are unsuitable for interpretation. The expert usually is not able to determine, why the model made a particular decision.

On contrary, the models, where the connection between the data samples is based on the physical principles, are potentially interpreted. For example, one can built the physical law in form of function \cite{schmidt2009distilling}, ordinary differential equations system \cite{brunton2016discovering,kondrashov2015data}. However, most of the physical laws are written in form of the partial differential equation (PDE).

Partial differential equations (PDE) are able to represent a vast variety of processes, which occur in dynamic systems in nature and society. For example, the Navier-Stokes equation, that describes the flow of liquids or gasses, is often used to solve hydrodynamic problems. Maxwell’s equations, which represent electromagnetic fields, is comprised of four partial differential equations, characterizing interactions between changes of electric and magnetic fields and the effects of charges and electric currents. 

However, derivation of these equations previously involved a primarily mathematical and physical approach to the problem, requiring a deep understanding of the phenomenon’s nature. This, consequently, reduces the ability to create models for systems that modern science has little to no knowledge about. For instance, there is no uniform theory that determines societal and economic dynamics. Furthermore, various unsolved problems can occur even is such explored studies as hydrodynamics and metocean science. 

Data-driven algorithms are a solution for cases of systems, that we lack knowledge about. Nevertheless, in most cases raw observational data are available. The data-driven algorithms bring the ability to build the model for dynamical systems from time-series of data, received from in-field or laboratory observations.  The development of the data-driven methodology of partial differential equations derivation, combined with recent advances in technologies of measurements and probing, brings new opportunities for studying of metocean dynamic systems. 

Sparse regression is considered to be the main tool for selection of the leading terms of the differential equations \cite{rudy2017data}. The applied regularization is based on the addition of the L1 norm of the calculated weights to the least-square expression. One of the most popular methods used in PDE discovery is the least absolute shrinkage and selection operator (LASSO). The main feature of LASSO is the ability to mutate the loss function. Zero weights are chosen for terms, that poorly fit the input data, and, therefore, identify the structure of the PDE.

Previously, the problem of the discovery of the differential equation structure has been developed in a number of papers. From derivation of systems of equations, defining physical laws, by means of symbolic regression \cite{gray1998nonlinear,winkler2005new} to study dynamic systems, that are represented by a system of partial differential equations \cite{bongard2007automated,schaeffer2017learning}. 

The methods of PDE derivation, used in previous papers, usually utilize regression over the set of the pre-determined terms, that are usually comprised of different polynomial combinations of derivatives and functions. This limitation provides only the discovery of equations, that have a corresponding structure. The method, presented in this paper is referred below as EPDE. It is based on a combination of sparse regression, that applies sparsity for the small set of potential terms. Sets, in turn, can have arbitrary form, obtained during the evolution process. Also, the proposed way of calculation of terms’ weights values includes the application of linear regression over the non-normalized data for selected terms.

The paper is organized as follows, Section \ref{sec2} describes the problem of data-driven PDE discovery in details. Also, in Section \ref{sec2} dataset for experiments is described. Section \ref{sec3} describes the data-driven PDE discovery algorithm based on evolutionary optimization. Section \ref{sec4} is dedicated to the analysis of algorithm precision, stability, and robustness. Section \ref{sec5} concludes the paper.

\section{Problem statement and data acquisition} \label{sec2}

The developed EPDE algorithm is aimed at the derivation of the dynamic systems governing equation by time series, containing information about the studied function (temperature, velocity, etc.). At first, the approach must be applied for test cases, including artificially created data, acquired from numerically solved equations. This simplification gives chances to exclude noise from data and, therefore, check the algorithm’s behavior independently from external conditions, such as faults of the measurement equipment. The ability to manually select parts of data and compare results gives opportunities for stability tests. Also, a noise of any magnitude can be added to data to investigate the reaction of the algorithm to it. 

In this work, the algorithm was tested on the wave equation, Burgers and Korteweg-de Vries equations. They were solved numerically with the application of a finite-difference scheme to approximate time and spatial derivatives. For instance, the Crank-Nicolson method was utilized to solve the Burgers equation

\begin{table}[h!]
\centering
\caption{Equations used for algorithm validation.}\label{tab1}
\begin{tabular}{|l|l|}
\hline
Name &  Equation \\
\hline
Burgers equation &  $\frac{\partial u}{\partial t}=-u\frac{\partial u}{\partial x}+\mu \frac{\partial^2 u}{\partial x^2}$ \\
The wave equation &  $\frac{\partial^2 u}{\partial t^2}=\frac{1}{c^2}\frac{\partial^2 u}{\partial x^2}$ \\
Korteweg-de Vries equation & $\frac{\partial u}{\partial t}+6 u \frac{\partial u}{\partial x}+\frac{\partial u^3}{\partial x^3}=0$ \\
\hline
\end{tabular}
\end{table}

From the acquired field of equation solution, its time and spatial derivatives are calculated in order to be utilized further in regression. These derivatives are calculated by the finite-difference method (Eq.~\ref{eq1}) due to its simplicity and sufficient quality on noise-free data and small time and space steps. 

\begin{equation}
\label{eq1}
   \begin{array}{cc}
   \frac{\partial u}{\partial t}=~u_t(x,t)=~\frac{u_{j}^{i+1}-u_{j}^{i-1}}{2\Delta t};~\frac{\partial u}{\partial x}=~u_x(x,t)=~\frac{u_{j+1}^{i}-u_{j-1}^{i}}{2\Delta x};  \\
   \frac{{{\partial }^{2}}u}{\partial {{t}^{2}}}=~\frac{u_{j}^{i+1}-2u_{j}^{i}+u_{j}^{i-1}}{\Delta {{t}^{2}}};~\frac{{{\partial }^{2}}u}{\partial {{x}^{2}}}=~\frac{u_{j+1}^{i}-2u_{j}^{i}+u_{j-1}^{i}}{\Delta {{x}^{2}}};  \\
   \frac{{{\partial }^{3}}u}{\partial {{x}^{3}}}=~\frac{u_{j+2}^{i}-2u_{j+1}^{i}+2u_{j-1}^{i}-u_{j-2}^{i}}{2\Delta {{x}^{3}}}  \\
   \end{array}
\end{equation}

In cases, where the additional noise will be added or measurements are used, to numerically differentiate the solution, complex methods should be used. Tikhonov regularization and other types of regression are recommended due to their ability to remove noise so that the PDE discovery algorithm obtain cleaned data as input.
After derivatives are obtained, it is possible to create vectors of spatial data for a specific time point:

\begin{equation}
\label{eq2}
   {{f}_{1}}\left( t \right)=\left[\begin{array}{cc}
1  \\
   \vdots   \\
   1  \\
   \vdots   \\
    1  \\
   \end{array}
   \right] \,,
      {{f}_{2}}\left( t \right)=\left[\begin{array}{cc}
   u\left( t,~{{x}_{0}} \right)  \\
   \vdots   \\
   u\left( t,~{{x}_{i}} \right)  \\
   \vdots   \\
   u\left( t,~{{x}_{n}} \right)  \\
   \end{array}
   \right]
   \,,
         {{f}_{3}}\left( t \right)=\left[\begin{array}{cc}
   {{u}_{x}}\left( t,~{{x}_{0}} \right)  \\
   \vdots   \\
   {{u}_{x}}\left( t,~{{x}_{i}} \right)  \\
   \vdots   \\
   {{u}_{x}}\left( t,~{{x}_{n}} \right)  \\
      \end{array}
   \right]
\, , ...
\end{equation}

After that, the normalization of each of these time frames should be held. It can be done with the highest variable value for that time point, or by time frame's L2-norm. Finally, data vectors are created by compositions of all time frames for the modeled period.
\begin{equation}
\label{eq3}
   E=\left[\begin{array}{cc}
{{f}_{1}^N}\left( t_1 \right)  \\
   \vdots   \\
{{f}_{1}^N}\left( t_j \right)  \\
   \vdots   \\
{{f}_{1}^N}\left( t_m \right)  \\
   \end{array}
   \right]=
   \left[\begin{array}{cc}
\frac{1}{\Vert{{f}_{1}}\left( t_1 \right)\Vert_2} \\
   \vdots   \\
\frac{1}{\Vert{{f}_{1}}\left( t_j \right)\Vert_2} \\\
   \vdots   \\
\frac{1}{\Vert{{f}_{1}}\left( t_m \right)\Vert_2} \\
   \end{array}
   \right]
   \,,
      U=\left[\begin{array}{cc}
{{f}_{2}^N}\left( t_1 \right)  \\
   \vdots   \\
{{f}_{2}^N}\left( t_j \right)  \\
   \vdots   \\
{{f}_{2}^N}\left( t_m \right)  \\
   \end{array}
   \right]=
   \left[\begin{array}{cc}
\frac{u(t_1,x_0)}{\Vert{{f}_{2}}\left( t_1 \right)\Vert_2} \\
   \vdots   \\
\frac{u(t_j,x_0)}{\Vert{{f}_{2}}\left( t_j \right)\Vert_2} \\\
   \vdots   \\
\frac{u(t_m,x_n)}{\Vert{{f}_{2}}\left( t_m \right)\Vert_2} \\
   \end{array}
   \right]
   \,, ...
\end{equation}

Where $f_j (t)$ is the L2-normalized vector from Eq.~\ref{eq2}.
    After normalized terms are found, the feature vectors F(j) are formed and written in terms Eq.~\ref{eq3} as, for example, the following product:
    
\begin{equation}
\label{eq4}
   F(j)=\left[\begin{array}{cc}
(u'(t_1,x_0) *u_t (t_1,x_0))^N \\
   \vdots   \\
(u'(t_m,x_n) *u_t (t_m,x_n))^N \\
   \end{array}
   \right]=
U_x*U_t
\end{equation}    

On a balance, the data preparation step consists in representing data and their spatial and time derivatives in vectors in form Eq.~\ref{eq3}. After these steps, features are collected in form Eq.~\ref{eq4} in order to perform the optimization procedure.

    \section{Algorithm description} \label{sec3}
    
\begin{figure}[h!]
\includegraphics[width=1\textwidth]{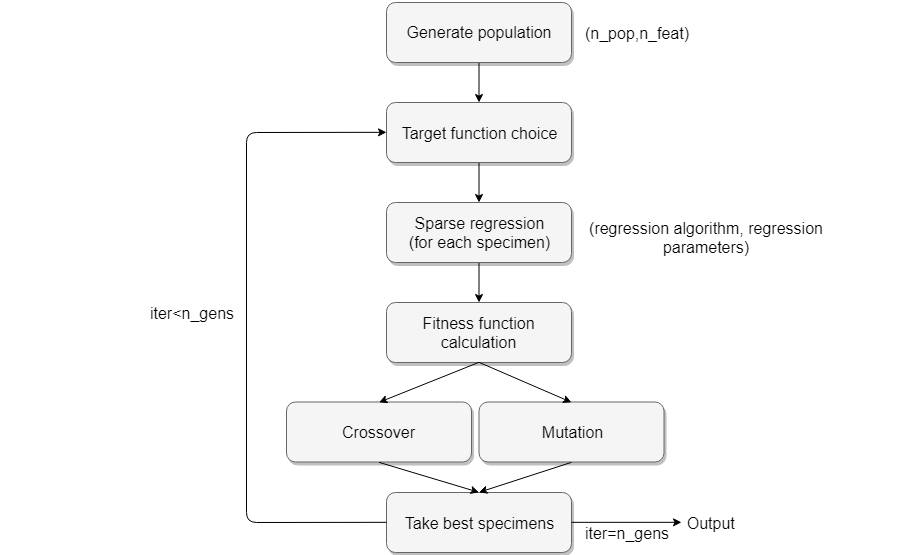}
\caption{Algorithm workflow scheme} \label{fig_algo_diag}
\end{figure}

The proposed algorithm includes two parts: the evolutionary algorithm that generates a small group of terms that are called individuals and sparse regression that allows choosing significant terms in the set of individuals. This process is shown schematically in Fig.\ref{fig_algo_diag}.

Such approach, in contrast to the existing algorithms, allows, on one hand, to generate more flexible space of terms for the sparse regression, and, on the other hand, to reduce the number of terms for the regression. Therefore, generation of the all possible terms is not required, as it is done within the evolutionary part. Sec. \ref{sec31} contains the description of the sparse regression part of the algorithm. Sec. \ref{sec32} contains the description of individual generation, crossover, and mutation.
    
\subsection{Sparse regression} \label{sec31}
Sparse regression is one of the most common instruments for discovering partial differential equations due to its ability to create the form of terms that would compose the resulting PDEs. It is based least absolute shrinkage and selection operator that is commonly used in machine learning to prevent over-fitting by filtering insignificant and redundant features by penalizing coefficients before them. It is based on the addition of L1-norm of unknown coefficients into a function (Eq.\ref{eq5}), used in the least-squares algorithm:

\begin{equation}
\label{eq5}
    Q=~\sum \limits_j\,\Vert w\cdot {{x}_{j}}-{{y}_{j}} \Vert _{2}^{2}+\lambda {\Vert{w} \Vert_{1}}~\to \min \limits_w
\end{equation}

In the sparse regression, w represents the searched sparse vector of weights between features, initialized by matrix x, and vector y is the learning target. $\lambda$ is a sparsity constant that is set before the learning process. 
To find values of the weights $\alpha$, that is representing the system’s PDE, it is possible to define the loss function (Eq.~\ref{eq6}) in the following way, using the defined set of features and target vectors, created in the previous section:

\begin{equation}
\label{eq6}
   \min \limits_\alpha\,\left( \sum \limits_{k=0}^p\,\Vert{{F}_{k}}\alpha -{{F}_{target,k}}\Vert_{2}^{2}+\lambda {\Vert{\alpha }\Vert_{1}} \right)
\end{equation}

Where p is the number of features selected for the regression algorithm.
This application of the regularized regression is not able to discover the true values of the weights due to the fact, that it uses normalized vectors of target and features. However, it is able to select leading ones with their sign. Due to the addition of L1-norm, the loss functions must be minimized, using optimization algorithms, that are able to work with non-differentiable functions, such as the subgradient method.

After the structure is found, the coefficients are defined with the non-normalized data, i.e. features are written back in the form (2) and regression is used to find the coefficients.
Usually, in regression all possible combinations \cite{schaeffer2017learning} of the feature vectors Eq.~\ref{eq4} are chosen for minimization problem Eq.~\ref{eq6}. Thus, the optimization problem complexity grows exponentially as the maximal order of the derivative increases. With the evolutionary algorithm, described below, one can use multiple reduced optimization problems instead of full regression on a complete terms library. In existing algorithms as a target feature usually, the highest-order time-derivative is chosen

\subsection{Evolutionary algorithm} \label{sec32}

The second element of the EPDE method is the evolutionary algorithm, that is aimed at the construction of the most complete set of terms. By its iterations, the evolutionary algorithm should be able to select and preserve the most appropriate elements of the resulting equation.  Therefore, the sparse regression is done on every iteration of the evolutionary algorithm for every candidate in the population with a random selection of target among the set of terms.  

To initiate the method, it is required to create a population of individuals, represented by chromosomes, where each gene represents a combination of functions and their derivatives. An evolutionary algorithm is able to vary the chromosomes in two ways: crossover, that represents the exchange of corresponding genes between two individuals, and mutation, which involves random alteration of chromosome’s genes. In the examined case, the mutation is held by the conversion of one term to the other randomly generated one. 

Due to the specification of the task, every individual represents a specific case of the equation, having its own features matrix and the target vector. Vectors F(i), that compose the columns of the feature matrix S (Eq.~\ref{eq7}), are created as a product of a randomly selected number of feature factors Eq.~\ref{eq4}:

\begin{equation}
\label{eq7}
S=\left[
    \begin{array}{cccc}
        | & | &| & \\
        F(1) & F(2) &F(3)& \, ... \\
        | & | &|& \\
    \end{array}
    \right]
\end{equation}

It should be emphasized, that the number of feature vectors in Eq.~\ref{eq7} is the parameter of the evolutionary algorithm. The second remark is that, in contrast to the existing algorithms \cite{rudy2017data,schaeffer2017learning,kondrashov2015data},  the target feature is chosen randomly, whereas in the sparse-regression only cases time-derivative is used.

While mutation is usually applied to all individuals of the population, crossover occurs only between the most eligible of them. To select candidates for crossover, the fitness function should be introduced. For the task of partial differential equation derivation, it can be introduced by a norm of the difference between the target term and the expression with other ones i.e. regression error, calculated for all of the training data:

\begin{equation}
\label{eq8}
f_{fitness}=\frac{1}{\Vert F \cdot \alpha-F_{target} \Vert_2}
\end{equation}

\begin{figure}
\includegraphics[width=\textwidth]{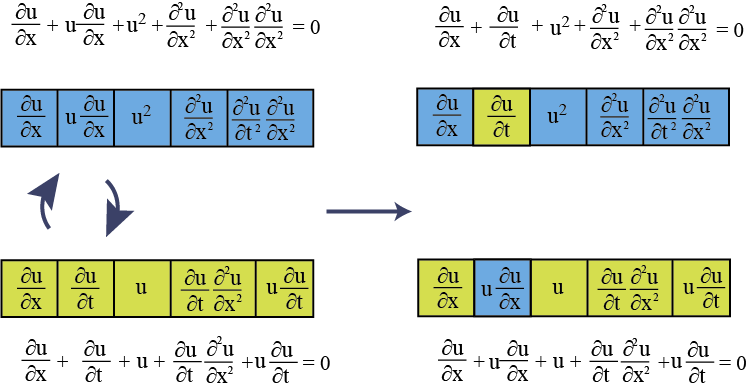}
\caption{An example of implemented crossover between two chromosomes, where each of them represent PDE} \label{fig1}
\end{figure}

A manner of the population’s participation in crossover should be defined before the initiation of an algorithm. In this research, the simplest way was adopted, executed by the selection of part the population part, that will breed. However, on some occasions that can result in stagnation of the evolution due to an occasional similarity between individuals with the highest fitness values, while the most eligible potential candidate requires gene from the fewer fit phenotypes. This limitation can be surpassed by the tournament selection. For that, a subset of the population shall be selected randomly and fitness tournaments must be held between them. The crossover procedure is schematically shown in Fig.~\ref{fig1}.

In this case, the required diversity of the population will remain and, consequently, there are higher chances of better individual’s creation. The individuals, generated during the crossover, replace the least fit ones in order to keep the population quantity stable while leaving the phenotypes, that previously had good values of fitness function intact. 
The iterative process can be stopped at the moment when the growth of the fitness function decreases below the defined threshold. Finally, the individual with the highest fitness at the last iteration is considered to be the one, that represents the structure of the partial differential equation, describing the dynamic system.

After the sparse regression application, one more regression step is required. It is initialized over the set of terms, selected by non-zero weights in the previous step. In this step, non-normalized fields of variables are used as a feature and target vectors. This 
approach is uncommon in general machine learning due to its limitations on variables of
different scale, where the algorithm is not able to properly generalize data and discover
a contribution of each feature. However, in this particular case, the structure of the
an equation, represented by weights of features, is already known, and these variables must
be evaluated according to their scale.

The described algorithm allows one to reduce regression space. Additionally, it allows to theoretically find ordinary differential equation instead of the PDE since target feature is not restricted by the highest time-derivative. This is required for potential one-dimensional static problems ODE discovery.

\begin{algorithm}[H]
 \KwData{matrix of variable values, time and space steps and ranges}
 \KwResult{target term, set of terms with their weights}
 Calculate derivatives of the variables\;
 Generate population of individuals, using defined individuals\_number and terms\_in\_individual\;
 \For{epoch = 1 to epoch\_number}{
  \For{individual in population}{
   Apply sparse regression to individual\;
   Calculate fitness function to individual\;
  }
  Hold tournament selection and crossover\;
  \For{individual in population}{
   Mutate individual\;
  }
  Select the individual with highest fitness function value\;
  Find true coefficients for discovered structure of the most fit individual;
 }
 \caption{The EPDE resulting pseudo-code}
\end{algorithm}

\section{Validation} \label{sec4}

To analyze the algorithm performance, it is necessary to make sure, that it has the following qualities: stability, approximation, and convergence. These qualities are dependent on each other, and to prove them, it is enough to check, if any two of them are fulfilled. Due to the reasons of convenience, in the research, stability, and convergence of the algorithm are studied. Convergence of the PDE deriving algorithm manifests in the improvement of the quality of the algorithm with the reduction of a step of the grid, from that is adopts data. Stability can be proved by addition of the noise to the input PDE solution and test, how this corruption affects the structure of the resulting equation.

The algorithm has proved to be capable of discovering partial differential equations structure and calculating the values of weights for the selected terms for all of the studied equations. 

At first, the algorithm was tested on the wave equation (Eq.~\ref{eq9}):

\begin{equation}
\label{eq9}
\frac{\partial^2 u}{\partial t^2}=\frac{1}{c^2}\frac{\partial^2 u}{\partial x^2}
\end{equation}

where $c = 2$ has the physical meaning of the sound velocity. It was solved on the spatio-temporal grid with 100x100 points.
Time and spatial steps were set as 0.1. The equation was solved by the implicit method. The resulting coefficients obtained by the evolutionary PDE discovery algorithm  are shown in Tab.\ref{tab_raw_coeffs}

\begin{table}
\centering
\caption{Resulting coefficients for the equations}\label{tab_raw_coeffs}
\begin{tabular}{|c|c|c|c|c|c|c|c|c|c|c|}
\hline
Term & 1 & u & $\frac{\partial u}{\partial t}$  & $\frac{\partial^2 u}{\partial t^2}$ & $\frac{\partial u}{\partial x}$ & $\frac{\partial^2 u}{\partial x^2}$ & $\frac{\partial^2 u}{\partial t^2} \frac{\partial^2 u}{\partial x^2}$ & $u\frac{\partial^2 u}{\partial t^2}$ &$u\frac{\partial u}{\partial x}$& $\frac{\partial^3 u}{\partial x^3}$\\
\hline
Wave Equation & 0 & 0 & 0 & 1 & 0 & -0.25006 & 0 & 0& 0& 0\\
\hline
Burgers & 0 & 0  & 1 & 0 & 0 & 0.09985 & 0 & 0 &-0.99986&0 \\
\hline
KdV & 0 & 0 & 1  & 0 & 0 & 0 & 0 & 0 & -5.9992&-0.9997\\
\hline
\end{tabular}
\end{table}

To make sure that the resulting structure does not change with the reduction of studied
space, the same algorithm was initiated over the parts of the matrix solution. For the
case of the wave equation, due to its relative simplicity, the same results have been
achieved on all studied sizes of the matrices: from the whole solution to 10$\%$ of it. These
results show that such simple structures of dynamic systems can be easily detected by the utilized algorithm.

The Burgers’ equation (Eq.~\ref{eq10}) presents an example of the more complex system to be studied:

\begin{equation}
\label{eq10}
\frac{\partial u}{\partial t}=-u\frac{\partial u}{\partial x}+\mu \frac{\partial^2 u}{\partial x^2}
\end{equation}

Where $\mu$ represents viscosity, which was set to a value of 0.1. The equation of solved on the grid of 256 x 256 spatial and time points correspondingly with steps of 16/256 and 10/256. To
acquire data for the algorithm from the equation solution, the Crank-Nicolson method was used. the resulting coefficients are shown in Tab.\ref{tab_raw_coeffs}

The selected part of the solution matrix has influence over the results of regressions
and, therefore, defines the equation’s structure. The results were tested on the parts of
the matrix from 1.0 to 0.1 of its size. On the lesser sizes of the selected matrix part,
especially for cases, when the selected part contains an only small part of the solution
ridge, the algorithm can have difficulties, deriving wrong structures. Consequently, the
incorrect set of terms prevents the calculation of their true weights during the second linear
regression phase. 

\begin{table}
\centering
\caption{Discovered structure of Burger`s and KdV equations for different input matrix section.}\label{tab_terms}
\begin{tabular}{|c|c|c|c|c|}
\hline
Data part & Burger`s correct & Burger`s wrong & KdV correct & KdV wrong\\
\hline
0.9 & $\frac{\partial u}{\partial t}$, $\frac{\partial^2 u}{\partial x^2}$, $u\frac{\partial^2 u}{\partial x}$ & - & $\frac{\partial u}{\partial t}$, $\frac{\partial^3 u}{\partial x^3}$, $u\frac{\partial^2 u}{\partial x}$& -\\
\hline
0.8 & $\frac{\partial u}{\partial t}$, $\frac{\partial^2 u}{\partial x^2}$, $u\frac{\partial^2 u}{\partial x}$ & - & $\frac{\partial u}{\partial t}$, $\frac{\partial^3 u}{\partial x^3}$, $u\frac{\partial^2 u}{\partial x}$& -\\
\hline
0.7 & $\frac{\partial u}{\partial t}$, $\frac{\partial^2 u}{\partial x^2}$, $u\frac{\partial^2 u}{\partial x}$  & - & $\frac{\partial u}{\partial t}$, $\frac{\partial^3 u}{\partial x^3}$, $u\frac{\partial^2 u}{\partial x}$& -\\
\hline
0.6 & $\frac{\partial u}{\partial t}$, $\frac{\partial^2 u}{\partial x^2}$, $u\frac{\partial^2 u}{\partial x}$ & - & $\frac{\partial u}{\partial t}$, $\frac{\partial^3 u}{\partial x^3}$, $u\frac{\partial^2 u}{\partial x}$ & -\\
\hline
0.5 & $\frac{\partial^2 u}{\partial x^2} $, $\frac{\partial u}{\partial t}$ & $\frac{\partial^2 u}{\partial x^2}\frac{\partial u}{\partial t}$& $\frac{\partial u}{\partial t}$, $\frac{\partial^3 u}{\partial x^3}$, $u\frac{\partial^2 u}{\partial x}$ & -\\
\hline
0.4 & $\frac{\partial u}{\partial t}$, $\frac{\partial^2 u}{\partial x^2}$, $u\frac{\partial^2 u}{\partial x}$ &- & $\frac{\partial u}{\partial t}$, $\frac{\partial^3 u}{\partial x^3}$, $u\frac{\partial^2 u}{\partial x}$& -\\
\hline
0.3  & $\frac{\partial^2 u}{\partial x^2} $, $\frac{\partial u}{\partial t}$ & $\frac{\partial^2 u}{\partial x^2}\frac{\partial u}{\partial t}$ & - & $\ \frac{\partial^2 u}{\partial x^2}$, $\frac{\partial^2 u}{\partial t^2}$\\
\hline
0.2  & $\frac{\partial^2 u}{\partial x^2} $, $\frac{\partial u}{\partial t}$ & $\frac{\partial^2 u}{\partial x^2}\frac{\partial u}{\partial t}$ &  -& $\ \frac{\partial^2 u}{\partial x^2}$, $\frac{\partial^2 u}{\partial t^2}$\\
\hline
0.1  & $\frac{\partial^2 u}{\partial x^2} $, $\frac{\partial u}{\partial t}$ & $\frac{\partial^2 u}{\partial x^2}\frac{\partial u}{\partial t}$ &  - & $\ \frac{\partial^2 u}{\partial x^2}$, $\frac{\partial^2 u}{\partial t^2}$\\
\hline
\end{tabular}
\end{table}

The results of the matrix division are presented in Table \ref{tab_terms}. Here it can be seen, that
proposed algorithm fails to discover the structure of the dynamic system in cases of the low
quantity of data: for sections, which are less than 0.5 of the solution and contain less
than 130 points, the algorithm tends to create wrong terms of the equation. Therefore,
to achieve the correct performance of the algorithm, data matrices of enough size shall be
passed to it. This condition creates some limitations for the method application for
cases of lack of data. 

Similar results have been attained for the Korteweg-de Vries equation. It has been
solved as shown in Table \ref{tab_raw_coeffs} - Table \ref{tab_raw_coeffs}. 
We note that for Table \ref{tab_raw_coeffs} different number of points was taken in order to check the performance of the algorithm. For Burger's equation, the 256x256 grid was taken whereas for the Korteweg-de Vries equation - 1024x1024 points.

The previously mention effect remained in this scenario: the algorithm only had issues in
discovering the structure of the governing equation. For cases, when it succeeded, the true values of the weights were calculated correctly, even on minor parts of the equation’s
solution matrix.

To check the evolutionary algorithm stability, the noise is added to the entire solution's field. It is added from a normally distributed random variable with zero mean value and dispersion taken as the fraction of maximal value.
As the invariant noise measure, Eq.\ref{eq11} is used.

\begin{equation}
\label{eq11}
    Q_{noise}=\frac{\Vert w_0-\widetilde{w} \Vert_2}{\Vert{w_0}\Vert_2}*100
\end{equation}

With $w_0$ in Eq.\ref{eq11} the initial (clean) solution field is designated, $\widetilde{w}$ is the field with noise added, $\Vert \cdot \Vert_2$ is the matrix's Frobenius norm.

For comparison, we take the latest supplementary code for the article \cite{rudy2017data} from GitHub repository. Same Burger`s equation solution field and same noise procedure implementation were taken. It should be noted, that we compare "basic" versions of the algorithms. For the sparse regression more sophisticated derivative procedure and meta-parameter optimization for the regression algorithm could be implemented, which, definitely, increases the quality of both algorithms.

As the discovery precision metric coefficient root mean square error is taken as it is shown in Eq.\ref{eq12}.

\begin{equation}
\label{eq12}
    E_{coeff}=\sqrt{\sum \limits_{i=1}^{N} (w_i-w_{pred})^2}
\end{equation}

In Eq. \ref{eq12}, $N$ is the number of terms taken for sparse regression, $w_i$ are the coefficients of clean (without noise) equation terms, $w_{pred}$ are the corresponding predicted coefficients.

\begin{figure}
\includegraphics[width=\textwidth]{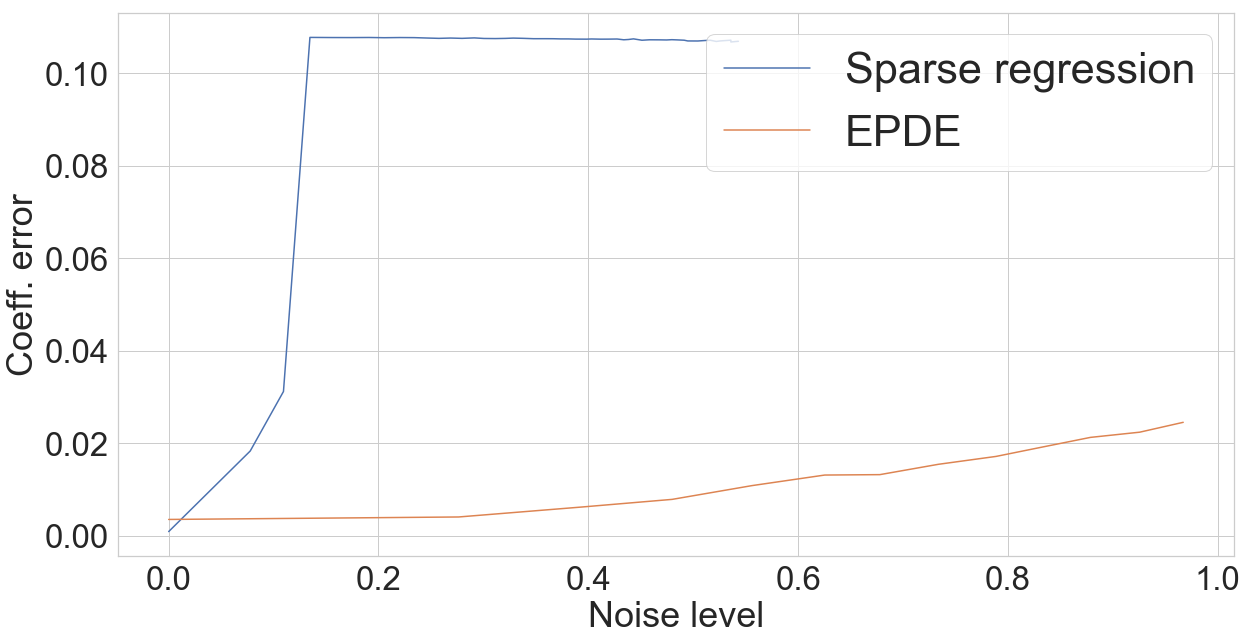}
\caption{An coefficient error with respect to the noise level Bruger's equation (blue - sparse regression \cite{rudy2017data}, orange - evolution algorithm)} \label{fig3}
\end{figure}

Polynomial derivatives procedure was utilized, also for the sparse regression improved ridge regression with $\alpha=10^{-6}$ was taken. Comparison results for the Bruger`s equation are shown in Fig.\ref{fig3}.
 
After certain noise level limit $Q_{noise} \approx 0.11$ (it is seen in Fig.\ref{fig3} as the function jump discontinuity), the classical algorithm loses ability to discover the term $\frac{\partial^2 u}{\partial x^2}$ without an additional regression tuning. However, it is still able to catch the leading term. In Fig.\ref{fig3} shown the maximal noise level range where the evolutionary approach is able to determine the equation structure.

We note that two other equations described in the article are not taken for the comparison due to the hyperbolic nature of the differential operators. Numerical solution of such equations requires specific methods that should be implemented for both algorithms. So, the comparison results, in this case, will show the quality of the implemented numerical schemes rather than the performance of the PDE discovery algorithms.

As seen the evolutionary approach allows one to extend the noise level which is allowed for all terms of the initial equation discovery. The term coefficients discovery precision is increased, which leads to more stable equation discovery and allows one to discover the equations in a more robust way.

\section{Conclusions and discussion} \label{sec5}
    
In the paper evolutionary approach for PDE discovery is described. In contrast to the existing algorithms based on the regression on a complete terms library it has the following advantages:
\begin{itemize}
    \item Regression is done on a reduced space, i.e. only a small amount of features is taken for the regression;
    \item More flexible features choice allows to obtain wider space of possible differential operators;
    \item No restriction on the target function is allowing to obtain more sophisticated forms of differential operators including ODEs;
\end{itemize}
    
The possible disadvantages could be:

\begin{itemize}
    \item Possible extended computation time due to the stochastic process of the initial population initialization, population crossover and mutation;
    \item Additional procedures are required in order to maintain the robustness of the algorithm, i.e. in order to obtain the same model for the data of the same origin;
\end{itemize}

The proposed method can be considered as a base point for the data-driven PDE discovery with an evolutionary approach. In the article, the main stages of the methods are shown. Every stage could be improved, for example, a more sophisticated grid function differentiation method could be taken to increase precision and stability. Also, more advanced evolution methods could be used in order to increase computation efficiency and stability.

The questions of the stability concerning the different boundary conditions, more difficult differential operators are left out of the scope of the paper. As well as the question of real observational data model discovery. However, the main goal was to show the proof-of-concept algorithm and validate it on synthetic examples and partially compare it with the existing algorithms.

%
%
%
\bibliographystyle{splncs04}
 \bibliography{mybibliography}
\end{document}